\documentclass[letterpaper, 10 pt, conference]{ieeeconf}  

\usepackage{graphics} 
\usepackage{amsmath} 
\usepackage{amssymb}  
\usepackage[colorlinks, linkcolor=blue, citecolor=blue]{hyperref}
\usepackage{cite}
\usepackage{graphicx}
\usepackage{caption}
\usepackage{subfigure}
\usepackage{mathrsfs}
\usepackage{array}
\usepackage{multirow}
\usepackage{makecell}
\usepackage{xcolor}
\usepackage{amsfonts}
\usepackage{setspace}
\usepackage{gensymb}
\usepackage{threeparttable}
\makeatletter
\newif\if@restonecol
\makeatother

\usepackage[linesnumbered,ruled,vlined]{algorithm2e}
\usepackage{algpseudocode}
\usepackage{amsmath}
\makeatletter
\renewcommand*\env@matrix[1][*\c@MaxMatrixCols c]{%
 \hskip -\arraycolsep
 \let\@ifnextchar\new@ifnextchar
 \array{#1}}
\makeatother

\IEEEoverridecommandlockouts                              

\title{\LARGE \bf
Globally optimal consensus maximization for robust visual inertial localization in point and line map
}

\author{Yanmei Jiao$^{1}$,
Yue Wang$^{1}$,
Bo Fu$^{1}$,
Qimeng Tan$^{2}$,
Lei Chen$^{2}$,
Minhang Wang$^{3}$,
\\Shoudong Huang$^{4}$ and Rong Xiong$^{1}$
\thanks{$^{1}$Yanmei Jiao, Yue Wang, Bo Fu, Rong Xiong are with the State Key Laboratory of Industrial Control and Technology, Zhejiang University, Hangzhou, P.R. China. $^{2}$Qimeng Tan and Lei Chen are with the Beijing Key Laboratory of Intelligent Space Robotic System Technology and Applications, Beijing Institute of Spacecraft System Engineering, Beijing, P.R. China. $^{3}$Minhang Wang is with the Application Lab, Huawei Incorporated Company, P.R. China. $^{4}$Shoudong Huang is with the Center for Autonomous Systems (CAS), the University of Technology, Sydney, Australia. Yue Wang is the corresponding author {\tt\small wangyue@iipc.zju.edu.cn}.}%
}

\begin{document}

\maketitle
\thispagestyle{empty}
\pagestyle{empty}

\begin{abstract}
Map based visual inertial localization is a crucial step to reduce the drift in state estimation of mobile robots. The underlying problem for localization is to estimate the pose from a set of 3D-2D feature correspondences, of which the main challenge is the presence of outliers, especially in changing environment. In this paper, we propose a robust solution based on efficient global optimization of the consensus maximization problem, which is insensitive to high percentage of outliers. We first introduce \emph{translation invariant measurements} (TIMs) for both points and lines to decouple the consensus maximization problem into rotation and translation subproblems, allowing for a two-stage solver with reduced search space. Then we show that (i) the rotation can be estimated by minimizing TIMs using only \emph{1-dimensional branch-and-bound} (BnB), (ii) the translation can be estimated by running 1-dimensional search for each of the three axes with \emph{prioritized progressive voting}. Compared with the popular randomized solver, our solver achieves deterministic global convergence without requiring an initial value. Furthermore, ours is exponentially faster compared with existing BnB based methods. Finally, our experiments on both simulation and real-world datasets demonstrate that the proposed method gives accurate pose estimation even in the presence of 90\% outliers (only 2 inliers).
\end{abstract}

\section{Introduction}

Visual inertial navigation system is popular for state estimation of mobile robots, autonomous vehicles and augmented reality applications. Many efforts have been paid to build accurate, consistent and efficient visual inertial odometry \cite{li2013high}\cite{leutenegger2015keyframe}\cite{forster2016manifold}. However, its inherent drift is unacceptable in long-term operation, requiring absolute pose estimation for correction. Map based visual inertial localization is therefore an important component in a complete navigation system, of which the goal is to estimate the absolute pose from a set of corresponding 2D image feature points and global 3D map points. In this problem, one main challenge is the robustness of the solver against outliers, i.e. incorrect feature correspondences. When high percentage of correspondences is outlier, the performance of the general pose estimator may be severely degenerated.


Pose estimation with outliers is often stated as a consensus maximization problem. One popular solution is random sample consensus (RANSAC), which has lots of variants \cite{fischler1981random}\cite{choi1997performance} and has been employed in many visual localization methods \cite{meer1991robust}\cite{hartley2003multiple}. The advantage of RANSAC is the simplicity for implementation, and the effectiveness in many scenarios with moderate percentage of outliers. However, RANSAC cannot tolerate extreme percentage of outliers, say 90\%. In addition, it cannot guarantee the deterministic global optimality due to the probabilistic convergence.

\begin{figure}[tp]
  \centering
  \includegraphics[width=0.5\textwidth]{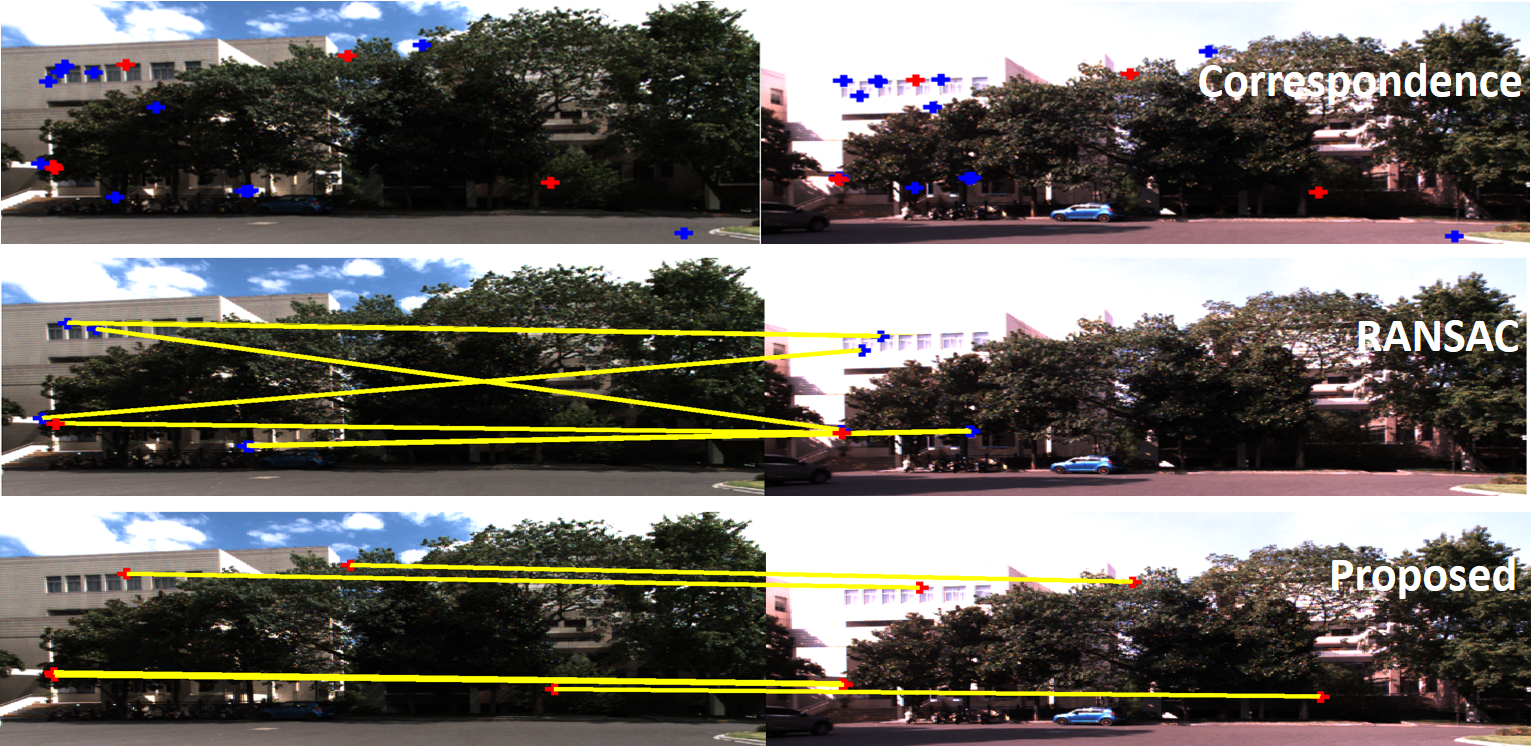}
  \caption{The projected map points on the map image (left column) and the detected image key points on the query image (right column), with inlier correspondences in red and outliers in blue. The initial correspondences found by feature descriptor matching (top), and the consensus set correspondences searched by RANSAC (middle) and proposed consensus maximization algorithm (bottom). }
  \label{fig.overview}
  \vspace{-0.5cm}
\end{figure}
In contrast to RANSAC, another solution to consensus maximization is global optimization based methods. It gives globally optimal solution without relying on an initial value \cite{brown2015globally}\cite{campbell2017globally}, while it cannot perform in real-time due to the considerable computation time. Most existing global optimization methods aim at general pose estimation problems. They employ branch-and-bound (BnB) as the basic framework to reduce the search space \cite{olsson2008branch}, or mixed integer programming for further acceleration \cite{li2009consensus}\cite{chin2016guaranteed}. But the computational cost is still unsatisfactory as the pose space $SE(3)$ is coupled. Even inertial measurement is provided, it cannot be easily substituted into the problem for decoupling.

In this paper, we propose a deterministic visual inertial localization solution to achieve global convergence with much higher efficiency. The key idea is to divide $SE(3)$ search space into multiple 1-D search spaces. Specifically, inspired by the decoupling idea in \cite{yang2019polynomial}, we build intermediate cost function for both point and line features, \emph{translation invariant measurements} (TIMs), to decouple consensus maximization into two cascaded subproblems only related to rotation $SO(3)$ and translation $\mathbb{R}^3$ respectively. Based on TIMs, the globally optimal rotation is then searched by \emph{1-dimensional BnB} in $[-\pi,\pi]$ with the aid of inertial measurements. For the translation, $\mathbb{R}^3$ search is replaced with three 1-dimensional $\mathbb{R}$ search for each axis using \emph{prioritized progressive voting}. To the best of our knowledge, this is the first solver for visual inertial localization with deterministic global optimality. In summary, our contributions include
\begin{itemize}
  \item TIMs based formulation of visual inertial localization that decouples the problem and enables 1D BnB based global optimization of the rotation.
  \item Prioritized progressive voting method that replaces $\mathbb{R}^3$ space search with three $\mathbb{R}$ search for global optimization of the translation.
  \item Experiments on simulation and real-world cross-session datasets that validate the effectiveness and efficiency of the proposed method against comparative methods.
\end{itemize}

The remainder of the paper is organized as follows: Section II reviews the related literatures. Section III presents the decoupling of the consensus maximization problem. Section IV introduces the solutions of the subproblems. Section V demonstrates the experimental settings and results, followed by Section VI concluding the paper.

\section{Related Works}

\subsection{Visual localization}

Visual localization and navigation for mobile robots has been studied extensively in the robotics and computer vision communities in the recent decade. A general visual navigation system has two components: visual odometry, which estimates the relative pose and has drift in long term \cite{nister2004visual}\cite{Geiger2011IV}, and visual localization, which eliminates the drift by registering the image on a global map \cite{furgale2010visual}\cite{tang2018topological}. More recently, inertial sensors are employed in the system to improve the accuracy and robustness \cite{mourikis2007multi}\cite{lynen2015get}\cite{schneider2018maplab}. Specifically, the inertial sensor has globally observable pitch and roll measurements, reducing the degrees of freedom (DoF) in visual inertial localization problem to 4. In \cite{li2013high}\cite{qin2018vins}, the reduction is utilized when formulating the pose estimation given a set of inlier feature correspondences. However, few works have been done on outliers elimination when inertial measurements are provided.

\subsection{Random sample consensus}

For robust localization given the feature correspondences containing outliers, RANSAC is the most popular solution employed in many visual navigation system. To deal with the visual localization without inertial measurements, i.e. 6DoF, there have been many variants. In \cite{gao2003complete}\cite{lepetit2009epnp}\cite{wang2018efficient}, point feature correspondences based RANSAC are studied. In \cite{dhome1989determination}\cite{chen1990pose}\cite{ramalingam2011pose}, RANSAC is extended to line features. When inertial measurements are provided, the DoF of the problem is reduced, which is utilized by RANSAC to improve the robustness in \cite{kneip2011robust}\cite{kukelova2010closed}, and extended to both point and line correspondences in \cite{jiao20192}. As RANSAC is developed on randomized sampling theory, it is simple to implement and has good performance on scenarios with moderate outliers. But its disadvantage is also obvious, including low tolerance against extreme outliers, local convergence and no guarantee of the optimality \cite{speciale2017consensus}.


\subsection{Outlier resistent estimator}

Another branch to reject outliers is to refer other forms of cost functions instead of the squared error \cite{mactavish2015all}. In \cite{zhou2016fast}, Geman-McClure cost function is utilized for 3D-3D registration, which is insensitive to outliers. In \cite{bosse2016robust}, M-estimators in several typical robotics problems are presented. Switchable cost function is employed to solve pose graph optimization with outlier loop closures \cite{sunderhauf2012switchable}\cite{lee2013robust}. A more compact solver for such cost function is dynamic covariance scaling which is introduced in \cite{agarwal2013robust}. More recently, in \cite{yang2020graduated}, several forms of robust cost functions are unified and solved using graduated non-convexity without an initial guess, which demonstrates good performance in 3D-3D registration, pose graph optimization, and is extended to non-minimal solver for shape reconstruction from an image in \cite{yang2020perfect}. Alternatively, in \cite{tzoumas2019outlier}, the outlier rejection is solved by adaptively removing the measurements with large errors, which is simple but show superior performance than RANSAC. These methods achieve deterministic convergence, while some of them offer certifiable optimality (or sub-optimality guarantees).

\subsection{Global optimization method}

Global optimization methods are proposed to achieve the global optimality and deterministic convergence. In this branch of literatures, Branch-and-Bound (BnB) is mostly used, which gradually prunes the solution space by coarse-to-fine division. In \cite{breuel2003implementation}, BnB is used to solve the 2D-2D registration problems. In \cite{olsson2008branch}, a general framework for point, line and plane features is proposed to solve 3D-3D registration via BnB. Integrated with mixed integer programming, the BnB optimization can converge faster \cite{li2009consensus}\cite{chin2016guaranteed}. In \cite{speciale2017consensus}, the linear matrix inequality constraints are introduced to mixed integer programming, resulting in a general-purpose faster BnB for all 2D-2D, 2D-3D and 3D-3D geometric vision problems. In the works mentioned above, the rotation is modeled as a rotation matrix with matrix level constraints. Thus it is unclear about the incorporation of inertial measurements. In addition, there are also specialized globally optimal algorithms focusing on one class of problem. In \cite{yang2015go}\cite{liu2018efficient}, pairs of features are used to decouple the 3D-3D registration. In \cite{yang2019polynomial}, TEASER is proposed to decoupled scaled 3D-3D registration, achieving a fast three-stage optimization. These works show that it is possible to have superior performance with \emph{specialized} algorithms rather than only the \emph{general-purpose} framework, even also accelerated.

In this paper, we follow the idea of specialized solver to bridge the gap of globally optimal deterministic solution for visual inertial localization, which is a robust 3D-2D pose estimation problem with inertial measurements. To the best of our knowledge, this is the first work to study this problem in the context of global optimality. We expect this solution to be accurate and efficient.

\section{Decoupling Translation and Rotation}

The underlying problem of visual inertial localization is the pose estimation from 3D-2D correspondences with outliers. Formally, given a set $\mathfrak{P}$ consisting of correspondences between 3D global points $p_i \in \mathbb{R}^3$ and 2D visual points $u_i \in \mathbb{R}^2$, they satisfy
\begin{equation}\label{3d2d}
  u_i = \pi(Rp_i+t,K) + o_i + e_i
\end{equation}
where $R \in SO(3)$ and $t \in \mathbb{R}^3$ is the camera pose to be estimated, $\pi$ is the camera projection function with known intrinsic parameters $K$, $|e_i| < n_i$ is assumed to be bounded random measurement noise, $o_i$ is zero for inlier while an arbitrary number for outlier. To deal with outliers, the robust pose estimation generally begins with consensus maximization problem as
\begin{eqnarray}\label{costrt}
  &\max_{R,t,\{z_i\}}\sum z_i \\
  &s.t.~~~~ z_i |u_i - \pi(Rp_i+t,K)| \leq n_i,~~i \in \mathfrak{P}
\end{eqnarray}
where $z_i$ is binary, indicating whether $o_i$ is zero. To solve the problem in global, general BnB algorithms search in $SE(3)$, which is a coupled space of $SO(3)$ and $\mathbb{R}^3$. But this probably leads to exponential computational complexity in bad cases. For local techniques like RANSAC, inliers may be estimated conservatively, i.e. inliers regarded as outliers, especially when the noise is unavoidable.

\subsection{Translation invariant measurements}

\subsubsection{Point-TIM}

Inspired by the minimal solution in RANSAC, we develop an intermediate measurement which is invariant to the translation of the pose. Mathematically, given an image key point $u_i$, we have an un-normalized direction vector from the camera center as
\begin{equation}\label{lineeq}
  \tilde{u}_i \triangleq \left(
    \begin{array}{c}
      \tilde{u}_{i,x} \\
      \tilde{u}_{i,y} \\
      1 \\
    \end{array}
  \right) = K^{-1}\left(
                    \begin{array}{c}
                      u_i \\
                      1 \\
                    \end{array}
                  \right)
\end{equation}
Then the corresponding world point $p_i$ is transformed to the camera coordinates and satisfies
\begin{equation}\label{linep}
  \frac{R_1 p_i + t_x}{\tilde{u}_{i,x}} = \frac{R_2 p_i + t_y}{\tilde{u}_{i,y}} = R_3 p_i + t_z
\end{equation}
where $R\triangleq (R_1^T,R_2^T,R_3^T)^T$ and $t \triangleq (t_x,t_y,t_z)^T$. Based on (\ref{linep}), we have two constraints from a correspondence. Naturally, given another correspondence $u_j$ and $p_j$, we can have two more constraints as
\begin{equation}\label{linep2}
  \frac{R_1 p_j + t_x}{\tilde{u}_{j,x}} = \frac{R_2 p_j + t_y}{\tilde{u}_{j,y}} = R_3 p_j + t_z
\end{equation}
According to (\ref{linep}) and (\ref{linep2}), we have linear constraints of the translation $t$. With proper variable substitutions among the constraints, and the globally observable pitch and roll angles from inertial measurements, we can eliminate $t$, reduce $SO(3)$ to $[-\pi,\pi]$, and derive TIM as
\begin{equation}\label{tim}
  d_p(\alpha) = d_{p,1} \sin \alpha + d_{p,2} \cos \alpha + d_{p,3}
\end{equation}
where $\alpha$ is the unknown yaw angle, $d_{p,1}$, $d_{p,2}$, $d_{p,3}$ and the derivation details are presented in the Appendix. Now we substitute the constraints which are related to both $R$ and $t$ in (\ref{costrt}) with the TIM, leading to
\begin{eqnarray}\label{costr}
  &\max_{R(\alpha),\{z_{ij}\}}\sum z_{ij} \\
  &s.t.~~~~ z_{ij} |d_{p,ij}(\alpha)| \leq n_{ij},~~i,j \in \mathfrak{P}
\end{eqnarray}
where $n_{ij}=\min(n_i,n_j)$, $z_{ij}=1$ indicates the $i$-th and $j$-th correspondence derived the constraint are inliers.

\subsubsection{Line-TIM}

Similar to a pair of point correspondences, given a set of line correspondences $\mathfrak{L}$, it is also possible to develop TIM. Given the end points of the image line segment $u_{k1}$ and $u_{k2}$, we have two un-normalized directions as (\ref{lineeq}), denoted as $\tilde{u}_{k1}$ and $\tilde{u}_{k2}$.

Then following the fact that the point $p_k$ on the world line lies on the plane spanned by the rays from camera center along direction $\tilde{u}_{k1}$ and $\tilde{u}_{k2}$, we have
\begin{equation}\label{sp}
  (\tilde{u}_{k1} \times \tilde{u}_{k2})^T(Rp_k+t) = 0
\end{equation}
which is a constraint for both rotation and translation. Since arbitrary number of points can be sampled from a line, we sample another point on the same world line to formulate the constraint as (\ref{sp}). Then only one line correspondence can lead to line-TIM after proper substitution as
\begin{equation}\label{timl}
  d_l(\alpha) = d_{l,1} \sin \alpha + d_{l,2} \cos \alpha + d_{l,3}
\end{equation}
where the line-TIM has the same form as point-TIM in (\ref{tim}), but the coefficients are different. The derivation details are also presented in the Appendix.

\textbf{TIMs based rotation only problem.} Note that either (\ref{tim}) or (\ref{timl}) is only related to the yaw angle. By combining them together, we have a general consensus maximization problem with TIM constraints only related to rotation compatible to the map having both point and line features as
\begin{eqnarray}\label{costrtim}
  &\max_{R(\alpha),\{z_{*}\}}\sum z_{*} \\
  &s.t.~~~~ z_{ij} |d_{p,ij}(\alpha)| \leq n_{ij},~~i,j \in \mathfrak{P}\\
  &~~~~~ z_k |d_{l,k}(\alpha)| \leq n_k,~~k \in \mathfrak{L}
\end{eqnarray}


\subsection{Two-stage consensus maximization solver}

With TIMs for both point and line correspondences, we decouple the original consensus maximization problem into rotation only problem, and translation only problem when the rotation is fixed. Accordingly, the proposed solver has two stages in cascade:
\begin{itemize}
  \item We estimate the rotation $\hat{R}$ by $R(\hat{\alpha})$ based on the TIMs in (\ref{costrtim}). This estimator solves a 1D optimization problem and is described in Section \ref{er}.
  \item We estimate the translation $\hat{t}$ based on the original consensus maximization in (\ref{costrt}) where the rotation is assigned with $\hat{R}$. This estimator solves a $\mathbb{R}^3$ optimization problem and is described in Section \ref{et}.
\end{itemize}

\section{Estimators of Rotation and Translation}

\subsection{BnB based optimization for rotation}\label{er}

We employ BnB strategy to solve problem (\ref{costrtim}). The cost function in (\ref{costrtim}) relates to $\alpha$ and $z_{*}$. But it is obvious that when $\alpha$ is determined, $\{z_{*}\}$ is simply derived by evaluating the constraints. So we denote the cost function as $E(\alpha)$ that is explained as the number of inliers given a yaw angle $\alpha$.

\textbf{Upper bound of cost function.} We then derive the upper bound of $E(\alpha)$ on the subset $\mathbb{A}$, denoted as $\overline{E}(\mathbb{A})$, where $\alpha \in \mathbb{A} \subseteq[-\pi,\pi]$. Recall (\ref{tim}) and (\ref{timl}), as the forms of point-TIM and line-TIM are the same, we denote them as $d(\alpha)$. The lower bound of $|d(\alpha)|$ on $\mathbb{A}$, denoted as $\underline{d}(\mathbb{A})$, is derived as
\begin{equation}\label{timlb}
  \underline{d}(\mathbb{A}) =  \min |a_1 \sin(\alpha + a_2) + d_3|
\end{equation}
where the derivation of the coefficients are introduced in Appendix. Note that $\underline{d}(\mathbb{A})$ can be solved analytically without any iterations. Now we formulate a consensus maximization problem as
\begin{eqnarray}\label{costlb}
  &\max_{R(\alpha),\{z_{*}\},\alpha \in \mathbb{A}}\sum z_{*} \\
  &s.t.~~~~ z_{ij} \underline{d}_{p,ij}(\mathbb{A}) \leq n_{ij},~~i,j \in \mathfrak{P}\\
  &~~~~~ z_k \underline{d}_{l,k}(\mathbb{A}) \leq n_{k},~~k \in \mathfrak{L}
\end{eqnarray}
where the problem is defined on $\mathbb{A}$, and the TIMs constraints are replaced with tight lower bounds, relaxing the constraints and yielding an optimistic estimation of $\hat{z}_{*}$. We then have
\begin{eqnarray}\label{costub}
  E(\alpha) \leq \overline{E}(\mathbb{A}) = \sum \hat{z}_{*},~~\alpha \in \mathbb{A}
\end{eqnarray}
as a tight upper bound. The equality exists when all constraints give the same $\alpha$ with $d_{p,ij}(\alpha) = \underline{d}_{p,ij}(\mathbb{A})$ and $d_{l,k}(\alpha) = \underline{d}_{l,k}(\mathbb{A})$, which is only possible when noise is free.

\textbf{Accelerate BnB optimization.} With (\ref{costrtim}-\ref{costub}), we have the BnB search for globally optimal rotation, of which the pseudo code is listed in Algorithm \ref{bnb1}. Note that the main idea of BnB is to prune the solution space $\mathbb{A}$ when its upper bound $\overline{E}(\mathbb{A})$ is smaller than the current best estimates $E^*$. Therefore, if we have a fast solution to initialize a good $E^*$, most solution spaces can be pruned at early stage, significantly improving the search efficiency. To implement this idea, we use RANSAC \cite{jiao20192} to generate a rough initial $E^*$. In addition, we introduce a heuristics to balance the global optimality and the efficiency. The best $M$ estimated $\alpha$ during RANSAC is utilized to initialize $M$ subsets among $[-\pi,\pi]$. Each subset centers at each estimated $\alpha$ with a width $w$. When $w$ is large, global optimality is emphasized and vice versa. Another implementation trick is to store the respective inliers when evaluating (\ref{costlb}) on each subset $\mathbb{A}$. When $\mathbb{A}$ is further divided into smaller subsets, only the stored inliers within $\mathbb{A}$ are evaluated, instead of all constraints, saving lots of computational cost. These techniques are all shown to accelerate the search in the experimental ablation study without drop of accuracy.
\begin{algorithm}
    \caption{Globally Optimal Rotation Search}
    \label{bnb1}
    \KwIn{3D-2D feature correspondences $\mathfrak{P}$, $\mathfrak{L}$}
    \KwOut{Optimal $\alpha^{*}$}
    Initialize partition of $[-\pi,\pi]$ into subsets $\{\mathbb{A}_{i}\}$.\\
    Initialize best estimation $E^{*}$, $\alpha^{*}$.\\
    Insert $\{\mathbb{A}_{i}\}$ into queue $q$.\\
    \While{$q$ is not empty}
    {
        Pop the first subset of $q$ as $\mathbb{A}$.\\
        Compute $\overline{E}(\mathbb{A})$ as (\ref{costlb}).\\
        \If{$\overline{E}(\mathbb{A}) > E^{*}$}
        {
            Assign center of $\mathbb{A}$ as $\alpha_{c}$.\\
            Compute $E(\alpha_{c})$ as (\ref{costrtim}).\\
            \If{$E(\alpha_{c})>E^{*}$}
            {
                Update $E^{*}\leftarrow E(\alpha_{c})$, $\alpha^{*}\leftarrow \alpha_{c}$.\\
            }
            Subdivide $\mathbb{A}$ into subsets and insert into $q$.\\
        }
    }
\end{algorithm}
\subsection{Prioritized progressive voting for translation}\label{et}
When $R(\hat{\alpha})$ is estimated, the co-linear and co-planar constraints (\ref{linep}) and (\ref{sp}) are all linear constraints for $t$. Thus we can transform the consensus maximization problem with point and line constraints as
\begin{eqnarray}\label{costlsys}
  &\max_{t,\{z_i\}}\sum z_i \\
  &s.t.~~~~ z_i |A_i t+b_i| \leq n_{i},~~i \in \mathfrak{P} \cup \mathfrak{L}
\end{eqnarray}
where $A_i \in \mathbb{R}^{1\times3}$ and $b_i \in \mathbb{R}$ are the coefficients for linear constraints derived from (\ref{linep}) or (\ref{sp}) with estimated $R(\hat{\alpha})$. However, this problem still has coupled constraints for $t$ so that $\mathbb{R}^3$ search is indispensable.

\begin{figure}[tp]
  \centering
  \includegraphics[width=0.5\textwidth]{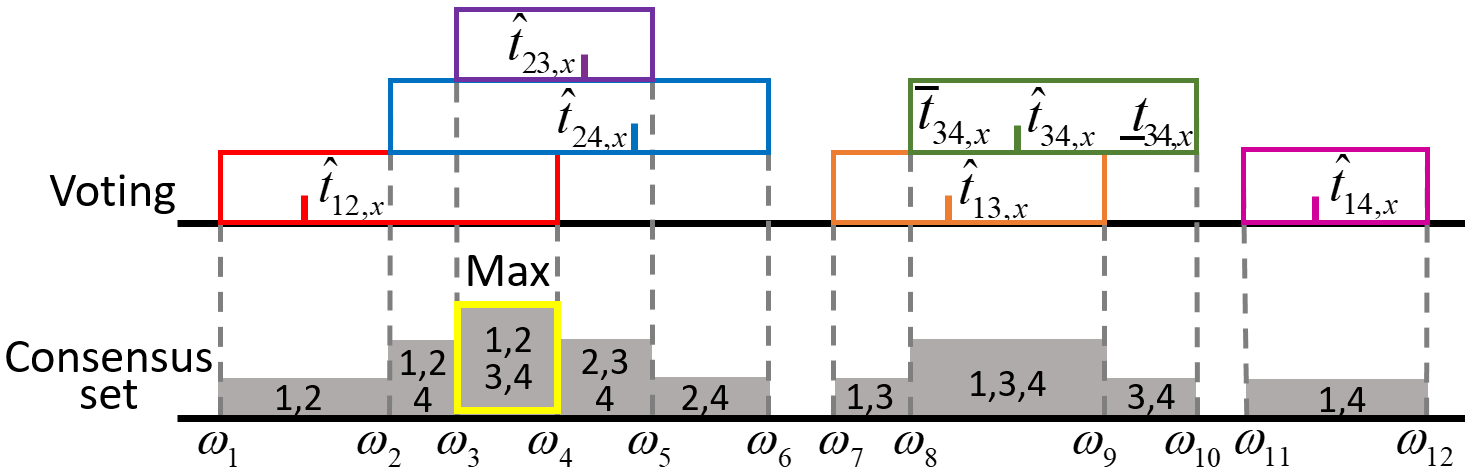}
  \caption{The voting illustration of $\hat{t}_x$. Each $\hat{t}_{ij,x}$ derived by $i$-th and $j$-th correspondence votes for the interval if $[\omega_{i},\omega_{i+1}]\subseteq[\underline{t}_{ij,x},\overline{t}_{ij,x}]$, which means the corresponding consensus set contains i and j.}
  \label{fig.vote}
  \vspace{-0.3cm}
\end{figure}

\textbf{Decoupled linear constraints.} Note that for a point correspondence constraint (\ref{linep}), we have two linear equations, while for a line correspondence constraint (\ref{sp}), we have one. Therefore, given a pair of correspondences including at least one point correspondence, say the $i$-th point correspondence and the $j$-th point or line correspondence, it is sufficient to solve $\hat{t}_{ij}$ for this small linear system (see Appendix for details), then we have
\begin{eqnarray}\label{costpsys}
  &\max_{t,\{z_{ij}\}}\sum z_{ij} \\
  &s.t.~~~~ z_{ij} |\hat{t}_{ij}-t| \leq n_{ij},~~i \in \mathfrak{P},j \in \mathfrak{P} \cup \mathfrak{L}
\end{eqnarray}
Now we find that the constraints are decoupled for each dimension of $t$. Set the $x$-dimension as example, we have
\begin{eqnarray}\label{cost1d}
  &\max_{t_x,\{z_{ij}\}}\sum z_{ij} \\
  &s.t.~~~~ z_{ij} |\hat{t}_{ij,x}-t_x| \leq n_{ij,x},~~i \in \mathfrak{P},j \in \mathfrak{P} \cup \mathfrak{L}
\end{eqnarray}
arriving at the resultant three dimension-wise linear constrained consensus maximization problems.

\textbf{Dimension-wise voting algorithm.} We use a voting algorithm to solve the problem. We first specify the noise bound $n_{ij,x}$ in (\ref{cost1d}). Given the noise bound $n_{i}$ in (\ref{costlsys}), we have the noise bound for $t$ following the techniques in \cite{mccormick1976computability} \cite{sherali1992new} as
\begin{equation}\label{tbound}
  \underline{t}_{ij} \leq \hat{t}_{ij} \leq \overline{t}_{ij}
\end{equation}
The details can be found in Appendix.

Still taking $x$-dimension as example, each estimated $\hat{t}_{ij,x}$ defines an interval $[\underline{t}_{ij,x},\overline{t}_{ij,x}]$. If the real $t_x$ lies in this interval, then the real inlier set contains the two correspondences deriving $\hat{t}_{ij,x}$. According to \cite{yang2019polynomial}, the insight is that the inlier set only changes its membership when real $t_x$ enters a new interval. Besides, given $K$ estimations, the \emph{maximum number of possible consensus sets, i.e. the cardinality of the solution space, is $2K-1$}, where $K$ is in \emph{quadratic} w.r.t the number of correspondences. This complexity enables a voting algorithm for all $2K-1$ sets. By counting the unique correspondences of the votes in each set, we get the corresponding consensus set. Then the maximal consensus set can lead to an estimation of $\hat{t}_x$. An illustrative case is shown in Fig. \ref{fig.vote} and the pseudo code is listed in Algorithm \ref{vt} with $x$-dimension as example. For simplicity, we replace $\hat{t}_{ij,x}$ with $\hat{t}_{k,x}$ in the pseudo code. Following the similar idea in \cite{yang2019polynomial}, by repeating the voting algorithm for three times, $\hat{t}$ is estimated as $[\hat{t}_x$, $\hat{t}_y$, $\hat{t}_z]^T$.

\begin{figure}[tbp]
  \centering
  \subfigure[]{
            \includegraphics[width=0.22\textwidth]{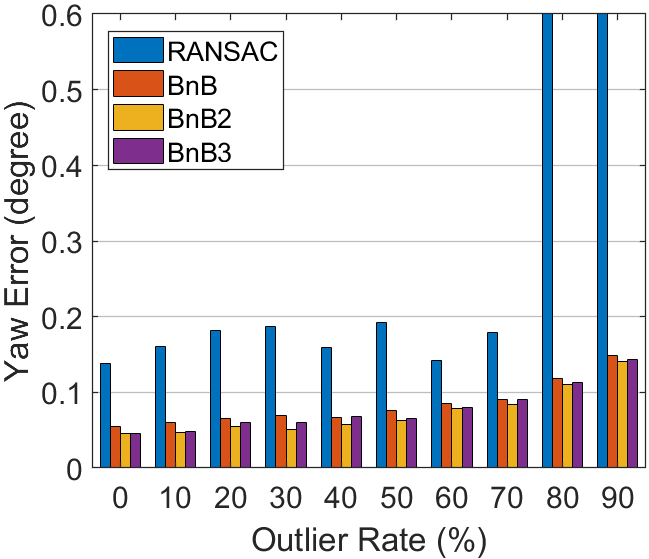}
        }
  \subfigure[]{
            \includegraphics[width=0.22\textwidth]{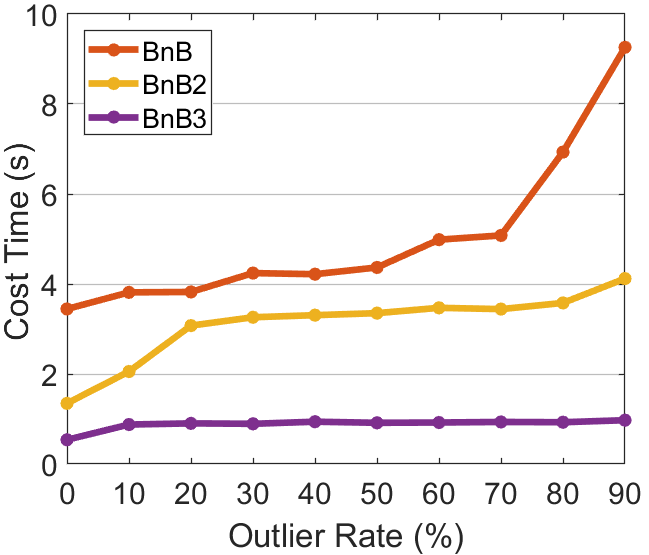}
        }
        \vspace{-0.25cm}
  \caption{The rotation accuracy and computation time over the increasing outlier rate. \emph{BnB2} denotes the BnB with RANSAC initialization. \emph{BnB3} denotes the BnB with both RANSAC initialization and the implementation trick.}
  \label{fig.yawTime}
        \vspace{-0.34cm}
\end{figure}

\begin{algorithm}
    \caption{Voting}
    \label{vt}
    \KwIn{$\{\hat{t}_{k,x}\}$, $\{\underline{t}_{k,x}\}$, $\{\overline{t}_{k,x}\},k=1..K$}
    \KwOut{Consensus sets $S$}
    Initialize key-value map $S$.\\
    $\omega = sort([ \underline{t}_{1,x},\overline{t}_{1,x},\underline{t}_{2,x},\overline{t}_{2,x},..,\underline{t}_{K,x},\overline{t}_{K,x} ])$.\\
    \For{$i=1..2K-1$}
    {
        $S([\omega_i,\omega_{i+1}])=\emptyset$.\\
        \For{$k=1..K$}
        {
            \If{$[\omega_{i},\omega_{i+1}]\subseteq[\underline{t}_{k,x},\overline{t}_{k,x}]$}
            {
                $S([\omega_i,\omega_{i+1}])=S([\omega_i,\omega_{i+1}])\cup k$.\\
            }
        }
    }
\end{algorithm}

\textbf{Prioritized progressive voting algorithm.} When the number of inliers is high, independent voting along three dimensions is possible. But when the number of inliers is low and outlier rate is high, independent dimension-wise voting may lead to failure. The reason is that, though it is almost impossible that there are more outliers than inliers having the similar $t$, \emph{it is possible that there are more outliers than inliers having the similar }$t_x$. In such scenario, search along $x$-dimension leads to incorrect $\hat{t}_x$, which cannot be corrected in the successive voting along $y$ or $z$-dimension.

To deal with such scenario while keeping a low computational complexity, we propose a prioritized progressive voting for translation in Algorithm \ref{voting}. The main idea is that we progressively vote on the three dimensions, but there is a priority, i.e. number of votes, for early termination. The experimental results show that the computational complexity of prioritized progressive voting is almost similar to the dimension-wise voting. Otherwise, it is also possible to use 3D BnB translation search for better accuracy, but it is slower because of the coupled multi-dimensional solution space. Finally, we apply nonlinear refinement to achieve the best accuracy when the maximum consensus set is found. \textcolor{blue}{Compared with the straightforward adaptive voting in \cite{yang2019polynomial} which only applicable when there are a certain number of inliers, the proposed prioritized progressive voting can deal with the situation with not only high outlier rate but also the low inlier number.}

\begin{figure}[tbp]
        \centering
        \subfigure[]{
            \includegraphics[width=0.225\textwidth]{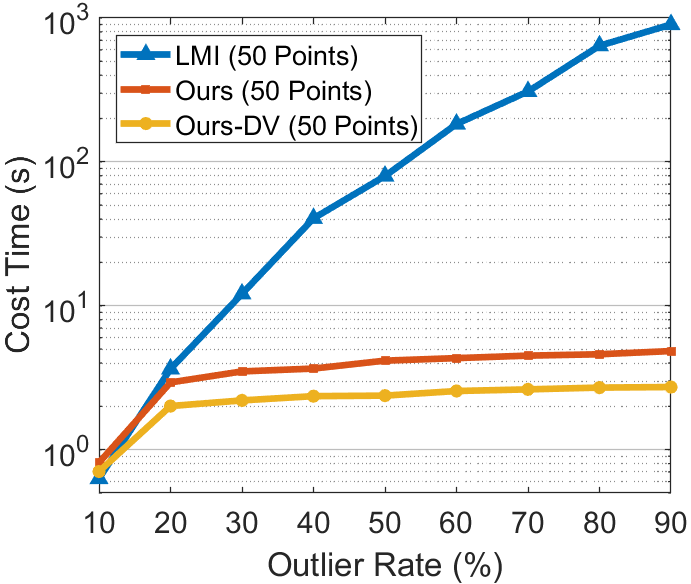}
        }
        \subfigure[]{
            \includegraphics[width=0.225\textwidth]{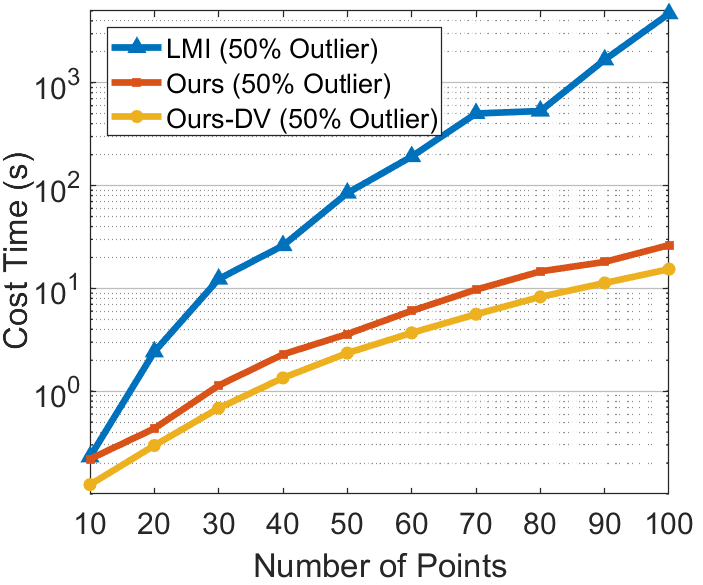}
        }
        \vspace{-0.25cm}
        \caption{Computation time comparison over increasing (a) outlier rate (b) number of points. \emph{Ours} denotes the proposed method with prioritized progressive voting, while \emph{Ours-DV} denotes the dimension-wise voting.}
        \label{fig.time}
  \vspace{-0.34cm}
\end{figure}

\begin{algorithm}
    \caption{Prioritized Progressive Voting}
    \label{voting}
    \KwIn{$\{\hat{t}_{k}\}$, $\{\underline{t}_{k}\}$, $\{\overline{t}_{k}\}$, $k=1..K$}
    \KwOut{Maximum consensus set $\hat{t}$}
    Initialize best estimation $E^{*}=0$.\\
    $S_x = Voting(\{\hat{t}_{k,x}\}, \{\underline{t}_{k,x}\}, \{\overline{t}_{k,x}\})$.\\
    Sort $S_x$ in decreasing cardinality.\\
    \For{each key $[i]$ in $S_x$}
    {
        \If{$|S_{x}([i])|<E^{*}$}
        {
            \textbf{break};\\
        }
        $S_y = Voting(\{\hat{t}_{k,y}\}, \{\underline{t}_{k,y}\}, \{\overline{t}_{k,y}\},k\in S_x([i]))$.\\
        \For{each key $[j]$ in $S_y$}
        {
            \If{$|S_{y}([j])|<E^{*}$}
            {
                \textbf{break};\\
            }
            $S_z = Voting(\{\hat{t}_{k,z}\}, \{\underline{t}_{k,z}\}, \{\overline{t}_{k,z}\},k\in S_y([j]))$.\\
            \If{$\max_{S_z([m])} |S_z([m])|>E^{*}$}
            {
                Update $E^{*} \leftarrow \max_{S_z([m])} |S_z([m])|$. \\
                Update $S^{*} \leftarrow \arg\max_{S_z([m])} |S_z([m])|$. \\
            }
        }
    }
\end{algorithm}

\section{Experimental Results}

In the experiments, we evaluate the proposed consensus maximization solver on (i) the feasibility and effectiveness of the subproblem solvers, (ii) the accuracy and robustness compared with existing methods, and (iii) the performance in real world visual inertial localization applications. We implement the proposed solver in MATLAB on a desktop with CPU Intel i7-7700 3.60GHz and 8G RAM.

\subsection{Ablation study}
We build the synthetic world consisting of 3D points and lines in the cube $[-1,1]^3$. The 2D image projections are generated with randomly sampled camera poses in $[-2,2]^3 \times [-\pi,\pi]^3$, as well as their inlier correspondences. All the projected 2D image points are added with bounded random noise $e_{i}$ with the bound $n_i=2$. Each outlier correspondence is generated from other randomly sampled camera pose different to ground truth pose. The total number of correspondences is fixed as 50. Specifically, there are 50 point correspondences when evaluating point only methods, while 25 point and 25 line correspondences for the point and line methods. We vary the outlier percentage from 10\% to 90\% with a step of 10\%. Statistic performance indicators are evaluated with an average of 100 Monte Carlo runs. Denoting the ground truth pose as $[R_{gt}|t_{gt}]$, we compute the translation error as $\triangle T=|\hat{t}-t_{gt}|$ in meter and the rotation error as the angle of $\triangle R=\hat{R}R_{gt}^{T}$ in degree.

\textbf{BnB heuristics.} We first evaluate the heuristics introduced in Section \ref{er} from the aspect of accuracy and efficiency. As shown in Fig. \ref{fig.yawTime}, with the heuristics, the efficiency is improved while the accuracy stays similar. Since the final pose is refined by nonlinear optimization, slight rotation error after BnB can be ignored. As a baseline, we also show the error of estimated rotation giving the most inliers in RANSAC, of which the performance is much worse, indicating inconsistency between the identified inliers and the real inliers. In following experiments, heuristics are applied with BnB as default setting.

\textbf{Translation voting.} We then compare the voting strategies introduced in Section \ref{et}. Now we can evaluate the final accuracy after nonlinear refinement. In addition to efficiency and accuracy, we also evaluate the consistency between the estimated consensus set and the real inlier set (CCI) using precision and recall. As shown in Fig. \ref{fig.time}, the computation of the prioritized progressive voting is slightly higher than the dimension-wise voting. More importantly, the increased time keeps almost consistent w.r.t outlier rate and correspondences number, which might be explained as no complexity growth for prioritized progressive voting. The CCI and accuracy are shown in the right columns in Tab. \ref{table.robustness}. We see that all variants achieve perfect CCI, naturally leading to high accuracy.

\begin{figure}[tp]
  \centering
  \includegraphics[width=0.45\textwidth]{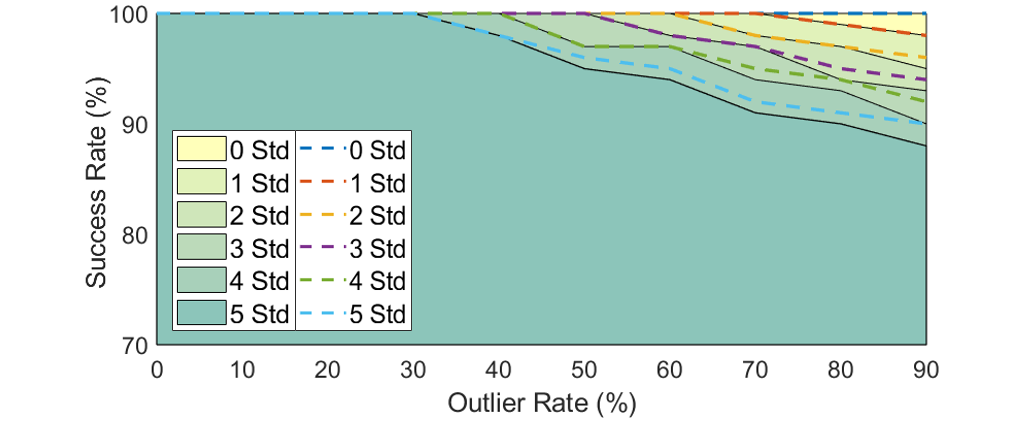}
  \caption{The sensitivity experiment result using proposed algorithm with dimension-wise voting (solid) and prioritized progressive voting (dash).}
  \label{fig.imu}
  \vspace{-0.3cm}
\end{figure}

\textbf{Sensitivity to noisy inertial measurements.} As inertial measurements are noisy, it is necessary to evaluate the sensitivity of the proposed method. We add Gaussian noise with zero mean and increasing standard deviation up to 5 degree on both pitch and roll angle. The threshold to judge a successful localization is 0.1m for translation error and 0.5 degree for rotation error as in \cite{miraldo2018minimal}. The result is shown in Fig.~\ref{fig.imu}, indicating that the proposed algorithm can achieve over 90\% success rate when the noise increases to 5 degree. This level of noise is far more than the pitch and roll estimations in practice \cite{bloesch2015robust}. In addition, we can find that the performance is better when employing prioritized progressive search.

\subsection{Comparison on synthetic datasets}
The comparative methods include the RANSAC-based methods EPnP\cite{lepetit2009epnp}, P3P\cite{gao2003complete}, 2-Entity\cite{jiao20192} and globally optimal method LMI\cite{speciale2017consensus}. We use the OpenCV\cite{itseez2015opencv} implementation of EPnP and P3P. For LMI, we modify their open source code in MATLAB following the paper, since only code for 3D-3D registration is released. In addition, we control the evaluation data having rotation angle less than $60\degree$ and add it as the constraint of LMI, as suggested in \cite{speciale2017consensus}. The 2-Entity RANSAC is implemented in MATLAB and we select the mixed sampling strategy which utilize both points and lines for pose estimation. All methods are followed by nonlinear refinement on the identified consesus set. We still use the synthetic dataset as in the ablation study.

\begin{figure}[tp]
  \centering
  \includegraphics[width=0.5\textwidth]{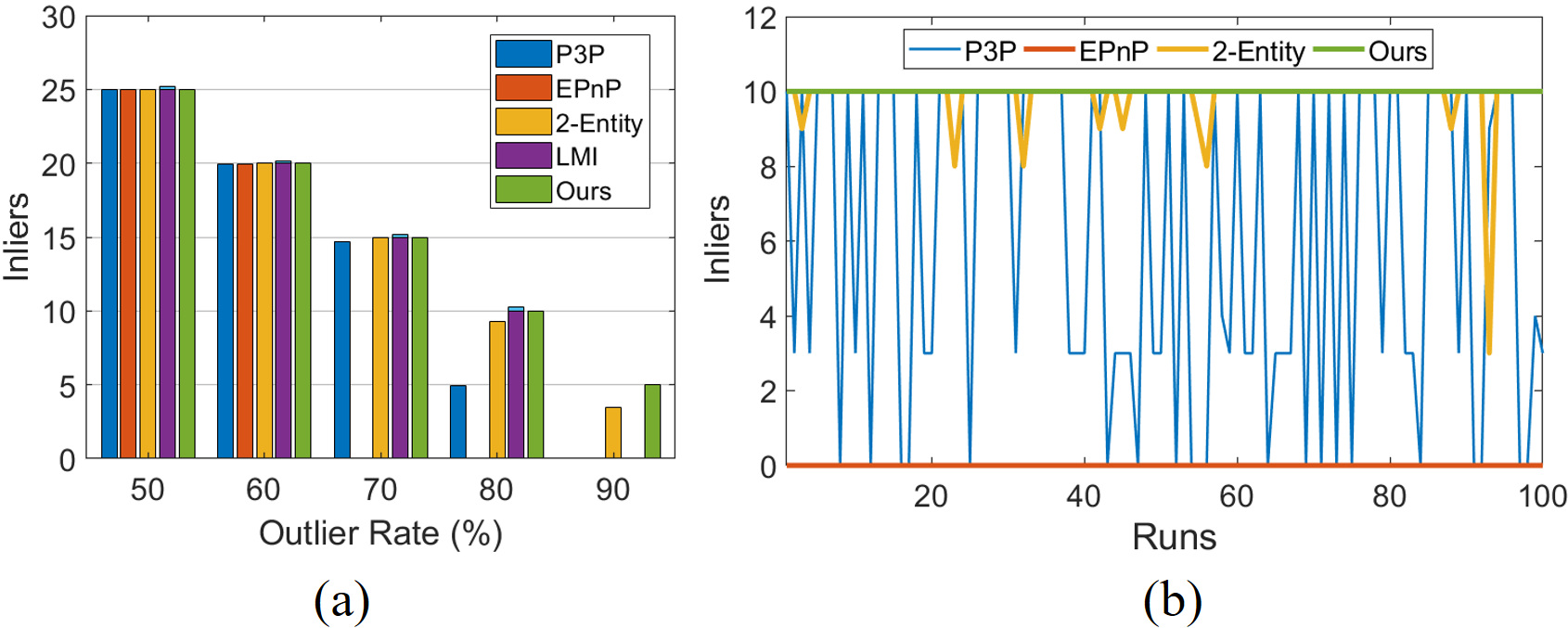}
  \vspace{-0.5cm}
  \caption{(a) The number of inliers in the estimated maximal consensus set w.r.t increasing outliers of successful estimation. (b) The number of inliers in the estimated maximal consensus set for 100 runs when the outlier rate is 80\%.}
  \label{fig.robustness}
  \vspace{-0.4cm}
\end{figure}

\begin{figure*}[htbp]
        \centering
        \subfigure[]{
            \includegraphics[width=0.3\textwidth]{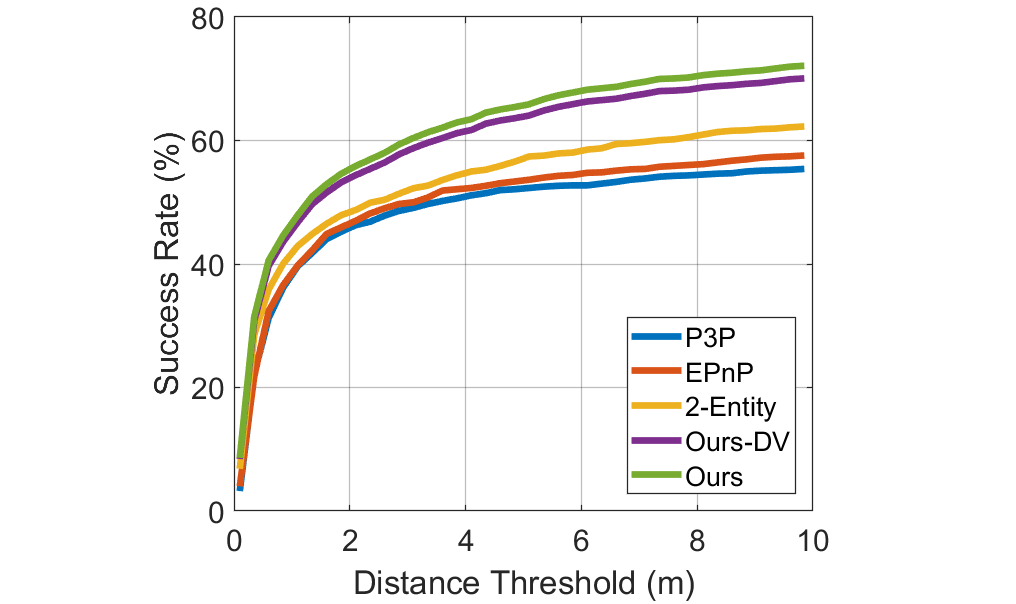}
        }
        \subfigure[]{
            \includegraphics[width=0.3\textwidth]{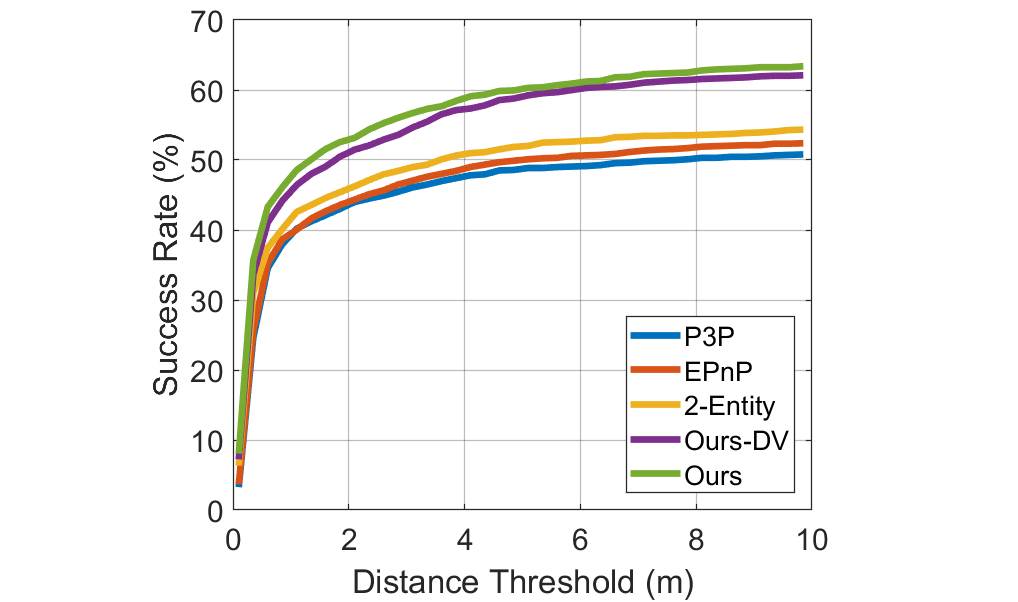}
        }
        \subfigure[]{
            \includegraphics[width=0.3\textwidth]{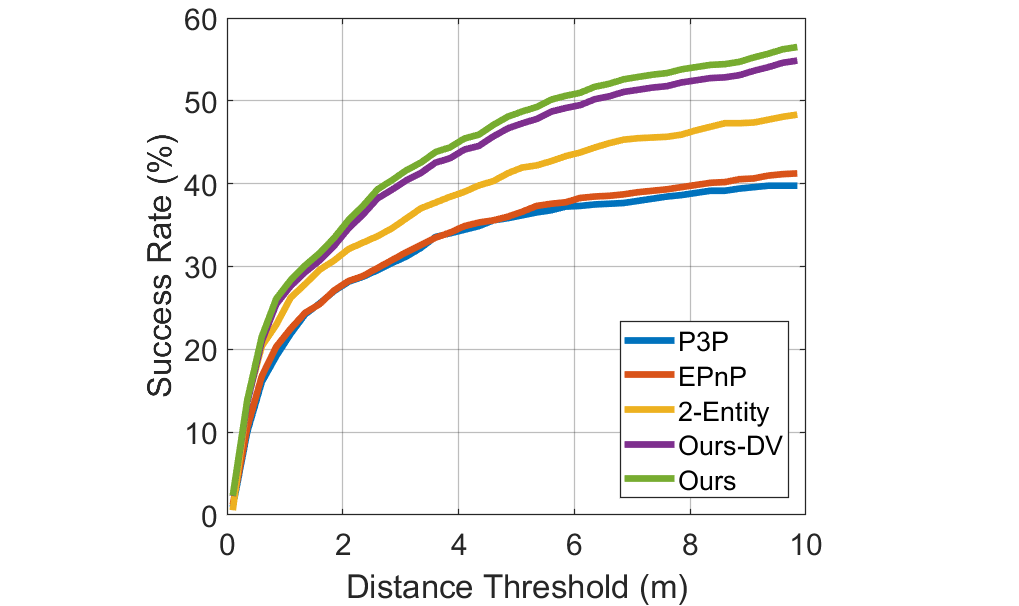}
        }
        \vspace{-0.3cm}
        \caption{Success rate with respect to threshold on the whole three sessions 0827 (left), 0828 (center) and 0129 (right).}
        \label{fig.session}
        \vspace{-0.4cm}
\end{figure*}

\textbf{Efficiency of globally optimal methods.} We first compare the efficiency between the proposed method and the LMI. We evaluate the computational cost with respect to the number of feature correspondences and the percentage of outliers. The result is shown in Fig. \ref{fig.time}, the computational cost of LMI is significantly higher than the proposed methods both for increasing number of correspondences, and the percentage of outliers. The growing gap may also indicate that the complexity of LMI is higher than ours.

\textbf{Deterministic convergence.} The vital difference between RANSAC and globally optimal method is the convergence. We compare the number of inliers in the estimated maximal consensus set with respect to increasing outliers when the final pose estimation is successful. The result is shown in Fig. \ref{fig.robustness}, which indicates that the proposed solution achieves deterministic perfect CCI, while RANSAC gives conservative estimations with less inliers and LMI finds optimistic estimations by incorrectly regarding outliers as inliers. In addition, both RANSAC and LMI fail when the outlier rate is 90\%. The results for all 100 runs when the outlier rate is 80\% are also shown in Fig. \ref{fig.robustness}. We can see that the proposed algorithm deterministically finds the globally optimal consensus, while RANSAC achieves global optimality probabilistically.

\textbf{Robustness and accuracy.} We finally show the performance of all methods on the synthetic data, including accuracy, precision and recall to measure the CCI, with respect to percentage of outliers ranging from 60\% to 90\%. Note that we only evaluate the accuracy for successful trials, since result on incorrectly identified consensus set can lead to very large error, disturbing the accuracy. The result in Tab. \ref{table.robustness} first confirms that CCI is highly related to the accuracy, validating the feasibility of maximizing consensus set. RANSAC gives consistent conservative estimations, as the precision remains at a higher level compared with the recall. For LMI, the estimation is prone to regard the outliers as inliers, thus the recall is higher compared with precision. Considering that LMI, P3P and EPnP are designed for general visual localization, the better performance achieved by 2-Entity and the proposed method, designed for visual inertial localization, is reasonable. But we can still summarize that superior result can be found by specialized globally optimal method.

\begin{table}[tbp]
\caption{Accuracy and CCI comparison.}
\vspace{-0.2cm}
\begin{center}
\begin{spacing}{1.0}
\resizebox{0.5\textwidth}{!}{
\begin{threeparttable}
\begin{tabular}{c l c c c c c c c c}
\Xhline{1.1pt}
 Outlier & Method & {P3P} & {EPnP} & {2-Entity} & {LMI} & {Ours-DV} & {Ours} \\
\hline
\multirow{4}{*}{60\%} & $\triangle$T(m) & 0.0010 & 0.0009 & 0.0008 & 0.0128 & \textbf{0.0005} & 0.0006 \\
& $\triangle$R(\degree) & 0.0196 & 0.0170 & 0.0059 & 0.0083 & \textbf{0.0019} & 0.0020 \\
& Precision & \textbf{1.00} & \textbf{1.00} & \textbf{1.00} & 0.96 & \textbf{1.00} & \textbf{1.00}\\
& Recall & 0.99 & 0.99 & \textbf{1.00} & {0.98} & \textbf{1.00} & \textbf{1.00}\\
& Success\% & \textbf{100} & \textbf{100} & \textbf{100} & 65 & \textbf{100} & \textbf{100}\\
\hline
\multirow{4}{*}{70\%} & $\triangle$T(m) & 0.0013 & - & 0.0011 & 0.0209 & \textbf{0.0005} & 0.0006\\
& $\triangle$R(\degree) & 0.0213 & - & 0.0211 & 0.1059 & \textbf{0.0017} & 0.0028\\
& Precision & \textbf{1.00} & 0 & \textbf{1.00} & 0.93 & \textbf{1.00} & \textbf{1.00}\\
& Recall & 0.98 & 0 & 0.99 & {0.93} & \textbf{1.00} & \textbf{1.00} \\
& Success\% & \textbf{100} & 0 & \textbf{100} & 54 & \textbf{100} & \textbf{100}\\
\hline
\multirow{4}{*}{80\%} & $\triangle$T(m) & 0.0017 & - & 0.0017 & 0.0246 & 0.0007 & \textbf{0.0006} \\
& $\triangle$R(\degree) & 0.0267 & - & 0.0257 & 0.4778 & 0.0050 & \textbf{0.0032}\\
& Precision & \textbf{1.00} & 0 & \textbf{1.00} & 0.46 & \textbf{1.00} & \textbf{1.00} \\
& Recall & 0.49 & 0 & 0.93 & {0.58} & \textbf{1.00} & \textbf{1.00} \\
& Success\% & 52 & 0 & 96 & 37 & \textbf{100} & \textbf{100}\\
\hline
\multirow{4}{*}{90\%} & $\triangle$T(m) & - & - & 0.0027 & - & \textbf{0.0007} & \textbf{0.0007} \\
& $\triangle$R(\degree) & - & - & 0.0411 & - & 0.0073 & \textbf{0.0043}\\
& Precision & 0 & 0 & \textbf{1.00} & 0.27 & \textbf{1.00} & \textbf{1.00} \\
& Recall & 0 & 0 & 0.70 & 0.35 & \textbf{1.00} & \textbf{1.00}\\
& Success\% & 0 & 0 & 86 & 0 & \textbf{100} & \textbf{100}\\
\Xhline{1.1pt}
\end{tabular}
\begin{tablenotes}
	\footnotesize
	\item[1] The accuracy is evaluated for successful trials, the precision and recall of CCI are for all test trails.
    \item[2] Ours-DV denotes the proposed method with dimension-wise voting.
\end{tablenotes}
\end{threeparttable}}
\end{spacing}
\end{center}
\label{table.robustness}
\vspace{-1.1cm}
\end{table}

\subsection{Comparison on visual inertial localization}
Finally, we evaluate all the methods on a real world cross-session visual inertial localization task. The dataset employed is YQ-dataset\cite{ding2018laser}. In the dataset, there are three sessions collected in summer 2017, denoted as 2017-0823, 2017-0827 and 2017-0828, and one session in winter 2018 after snow denoted as 2018-0129. The 3D map is built with 2017-0823 session and the other three sessions are used to evaluate the localization performance, indicating the changing environment. The details to obtain the 3D-2D point and line correspondences can be found in Appendix. For evaluation, we compute the ground truth relative pose between the query camera and the map by aligning the synchronized LiDAR scans. For the pitch and roll angle, we use the estimation of visual inertial odometry \cite{mur2017visual}.

\begin{table}[tbp]
\caption{Performance on selected cases in real world.}
\vspace{-0.2cm}
\begin{center}
\begin{spacing}{1.0}
\resizebox{0.46\textwidth}{!}{
\begin{threeparttable}
\begin{tabular}{l c c c | c c c }
\Xhline{1.1pt}
 &{ExpID} & {$|\zeta_P|/N_P$} & {$|\zeta_L|/N_L$} & {ExpID} & {$|\zeta_P|/N_P$} & {$|\zeta_L|/N_L$} \\
 & 01 & 9/18 & 0/0 & 02 & 15/39 & 0/0 \\
 \hline
 \makecell[c]{Method \\} & \makecell[c]{$\triangle T$ \\(m)} & \makecell[c]{$\triangle R$ \\(\degree)} & \makecell[c]{Inliers$^{1}$ \\$|\zeta^{*}|/|\zeta|$} & \makecell[c]{$\triangle T$ \\(m)} & \makecell[c]{$\triangle R$ \\(\degree)} & \makecell[c]{Inliers$^{1}$ \\$|\zeta^{*}|/|\zeta|$} \\
 \hline
 EPnP & 0.9938 & 0.8025 & 7/12 & 0.9026 & 1.3255 & 11/21 \\
 P3P & 0.8187 & 0.6302 & 7/11  & 1.9751 & 0.5977 & 10/20 \\
 2-Entity & 0.6683 & 0.4351 & \textcolor{blue}{8}/10 & 0.5703 & 0.3378 & 12/21 \\
 LMI & \textcolor{blue}{0.1630} & \textcolor{blue}{0.1951} & \textcolor{red}{9}/13 & 0.2832 & 0.2155 & \textcolor{blue}{14}/19\\
 Ours-DV & \textcolor{red}{0.1207} & \textcolor{red}{0.1321} & \textcolor{red}{9}/\textcolor{white}{0}9 & \textcolor{blue}{0.1803} & \textcolor{blue}{0.1550} & \textcolor{blue}{14}/14 \\
 Ours & \textcolor{red}{0.1207} & \textcolor{red}{0.1321} & \textcolor{red}{9}/\textcolor{white}{0}9 & \textcolor{red}{0.1753} & \textcolor{red}{0.1334} & \textcolor{red}{15}/15 \\
 \hline
 &{ExpID} & {$|\zeta_P|/N_P$} & {$|\zeta_L|/N_L$} & {ExpID} & {$|\zeta_P|/N_P$} & {$|\zeta_L|/N_L$} \\
 & 03 & 21/65 & 0/2 & 04 & 23/48 & 7/15 \\
 \hline
 EPnP & 0.4506 & 0.9741 & 10/29 & 0.5504 & 0.7823 & 19/28\\
 P3P & 0.3213 & 0.8807 & 13/27 & 0.3678 & 0.4066 & 19/27 \\
 2-Entity & 0.3138 & 0.4603 & 15/27 & 0.1405 & 0.2055 & 27/33\\
 LMI & 0.2998 & 0.3786 & \textcolor{blue}{19}/44 & 0.2834 & 0.1769 & 22/28 \\
 Ours-DV & \textcolor{blue}{0.1407} & \textcolor{blue}{0.1743} & 21/23 & \textcolor{blue}{0.0309} & \textcolor{blue}{0.1607} & \textcolor{blue}{28}/29 \\
 Ours & \textcolor{red}{0.1382} & \textcolor{red}{0.1707} & \textcolor{red}{21}/23 &
  \textcolor{red}{0.0253} & \textcolor{red}{0.1509} & \textcolor{red}{30}/30 \\
\hline
 &{ExpID} & {$|\zeta_P|/N_P$} & {$|\zeta_L|/N_L$} & {ExpID} & {$|\zeta_P|/N_P$} & {$|\zeta_L|/N_L$} \\
 & 05 & 21/38 & 8/13 & 06 & 96/134 & 3/4 \\
 \hline
 EPnP & 1.0876 & 0.8111 & 13/25 & 0.2705 & 0.5202 & 93/112\\
 P3P & 1.0876 & 0.8111 & 13/25 & 0.1682 & 0.5243 & 90/98\\
 2-Entity & \textcolor{blue}{0.1732} & \textcolor{blue}{0.2687} & \textcolor{blue}{27}/29 & 0.1163 & 0.4623 & 95/108\\
 LMI & 0.7641 & 0.6394 & 16/28 & \textcolor{blue}{0.0891} & \textcolor{blue}{0.2812} & \textcolor{blue}{96}/102\\
 Ours-DV & \textcolor{red}{0.1671} & \textcolor{red}{0.1072} & \textcolor{red}{29}/29 & \textcolor{red}{0.0861} & \textcolor{red}{0.2791} & \textcolor{red}{99}/99\\
 Ours & \textcolor{red}{0.1671} & \textcolor{red}{0.1072} & \textcolor{red}{29}/29 & \textcolor{red}{0.0861} & \textcolor{red}{0.2791} & \textcolor{red}{99}/99\\
\Xhline{1.1pt}
\end{tabular}
\begin{tablenotes}
	\footnotesize
	\item[1] $|\zeta|$ denotes the number of identified inliers, while $|\zeta^{*}|$ the true inliers.
\end{tablenotes}
\end{threeparttable}}
\end{spacing}
\end{center}
\label{table.real}
\vspace{-1.2cm}
\end{table}

\textbf{Selected cases performance.} We first select several typical examples for evaluation as in \cite{speciale2017consensus} and the results are shown in Tab.~\ref{table.real}. The Exp01, Exp02 and Exp03 are cases with pure point features where Exp03 has lines as disturbance and the outlier rate in these three cases are all more than 50\%. The RANSAC-based methods perform poorly compared with the global optimization methods. One thing to note is that in real world dataset, dimension-wise voting brings slight performance drop, but still achieves superior performance against comparative methods. Also note that in Exp03, the proposed method gives optimistic results by regarding 2 outliers as inliers, which may be caused by unknown noise bound thus inappropriate threshold in real world data. In Exp04, Exp05 and Exp06, the utilization of good line features promotes the performance of point line methods obviously (2-Entity and ours). Overall, the results still confirm the conclusions in simulation.

\textbf{Full dataset performance.} Finally, we arrive at the success rate on the whole three sessions as shown in Fig.~\ref{fig.session}. As LMI is too slow to finish all the dataset, here we only show the result of ours and RANSAC methods. We first see that the proposed globally optimal methods consistently outperform the RANSAC methods on all three sessions. The other fact is that progressive prioritized voting brings the best accuracy over the one with dimension-wise voting, because of the consideration on extremely low number of inliers.

\section{Conclusions}

In this paper, we propose a robust solver designed for visual inertial localization problem, achieving global optimization of the consensus maximization problem with deterministic convergence, even when the percentage of outliers is very high, say 90\%. The key step in our solver is the derivation of \emph{translation invariant measurements} for both points and lines, thus decoupling the problem into two smaller subproblems. Then we propose 1D BnB and prioritized progressive voting to find globally optimal rotation and translation respectively, accelerating the search efficiency. The effectiveness of the proposed method is validated on both synthetic and real world dataset.

\begin{appendices}

\section{Derivation of TIMs}

With the aid of inertial measurements, the pitch and roll angle between the current query camera frame and the gravity-aligned world reference frame are globally observable, such that the rotation estimation of the query camera with respect to the world can be formulated as
\begin{align}\label{eq.Rwc}
{R_{{\mathcal{W}}{\mathcal{C}}}}&=R_{z}(\alpha)R_{y}(\check{\beta})R_{x}(\check{\gamma})\nonumber\\
&=\begin{bmatrix}
    c\alpha & -s\alpha & 0 \\
    s\alpha & c\alpha & 0 \\
    0 & 0 & 1
  \end{bmatrix}
  \begin{bmatrix}
    c\check{\beta} & 0 & s\check{\beta} \\
    0 & 1 & 0 \\
    -s\check{\beta} & 0 & c\check{\beta}
  \end{bmatrix}
  \begin{bmatrix}
    1 & 0 & 0 \\
    0 & c\check{\gamma} & -s\check{\gamma} \\
    0 & s\check{\gamma} & c\check{\gamma}
  \end{bmatrix}\nonumber\\
  &\triangleq\begin{bmatrix}
               a_{11} c\alpha+b_{11} s\alpha & a_{12} c\alpha+b_{12} s\alpha  & b_{13} s\alpha \\
               a_{21} c\alpha+b_{21} s\alpha  & a_{22} c\alpha+b_{22} s\alpha & a_{23} c\alpha\\
               a_{31} & a_{32} & a_{33}
             \end{bmatrix}
\end{align}
where $\check{\beta}$ and $\check{\gamma}$ denote the observed pitch and roll angle provided by inertial measurements, $\alpha$ denotes the yaw angle to be estimated, $\sin \alpha \triangleq s\alpha$, $\cos \alpha \triangleq c\alpha$. Therefore, the rotation matrix is only determined by the estimation of yaw, which is the same in ${R}$, as $R=R_{{\mathcal{W}}{\mathcal{C}}}^{T}$. Thus the degrees of freedom (DoF) of the rotation matrix estimation can be reduced to 1 with the aid of inertial measurements, that is
\begin{equation}\label{eq.R}
R=R(\alpha)=\begin{bmatrix}
               a_{11} c\alpha+b_{11} s\alpha & a_{21} c\alpha+b_{21} s\alpha & a_{31} \\
               a_{12} c\alpha+b_{12} s\alpha  & a_{22} c\alpha+b_{22} s\alpha & a_{32} \\
               b_{13} s\alpha & a_{23} c\alpha & a_{33}
             \end{bmatrix}
\end{equation}

\subsection{Derivation of point-TIM}

\begin{figure}[tbp]
    \centering
    \includegraphics[width=0.4\textwidth]{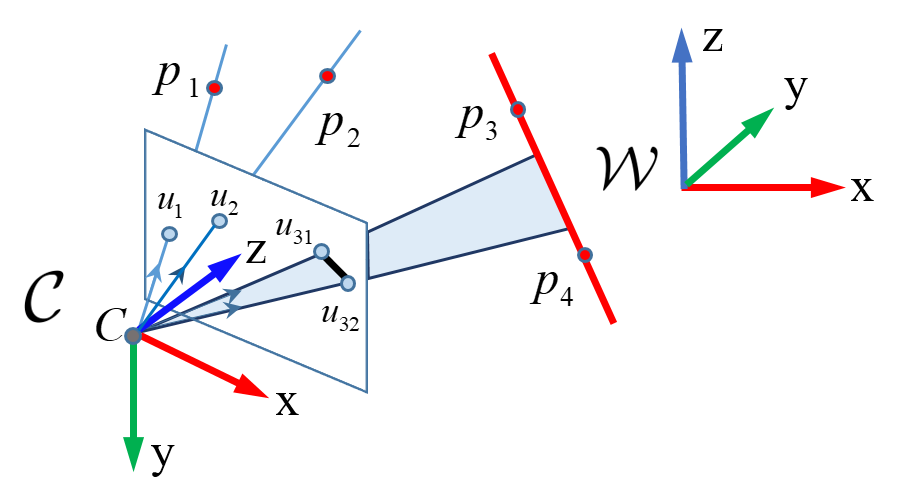}
    \caption{The illustration of 2D-3D point and line features.}
    \label{fig.coordinates}
\end{figure}

The collinearity of each 2D-3D point features is utilized to derive the point-TIM as shown in Fig.~\ref{fig.coordinates}. Mathematically, given an image key point $u_i$, we have an un-normalized direction vector from the camera center as
\begin{equation}\label{lineeq}
  \tilde{u}_i \triangleq \left(
    \begin{array}{c}
      \tilde{u}_{i,x} \\
      \tilde{u}_{i,y} \\
      1 \\
    \end{array}
  \right) = K^{-1}\left(
                    \begin{array}{c}
                      u_i \\
                      1 \\
                    \end{array}
                  \right)
\end{equation}

According to the projection geometry, the optical center of camera frame $C={\bf{0}}_{3\times1}$, the 2D point $\tilde{u}_{1}$ and the corresponding 3D point $p_{1}$ lie on the same line, which is denoted as $\{C,\tilde{u}_{1},{Rp_{1}+t}\}_{L}$. By solving the line equation from the first two points and substituting the third point into the equation, we have
\begin{equation}\label{linep}
  \frac{R_1 p_1 + t_x}{\tilde{u}_{1,x}} = \frac{R_2 p_1 + t_y}{\tilde{u}_{1,y}} = R_3 p_1 + t_z
\end{equation}
where $R\triangleq (R_1^T,R_2^T,R_3^T)^T$ and $t \triangleq (t_x,t_y,t_z)^T$. Based on (\ref{linep}), we have two constraints from a correspondence as
\begin{equation}\label{p1constraint1}
  \tilde{u}_{1,x}(R_{2}p_{1}+t_{y})-u_{1,y}(R_{1}p_{1}+t_{x})=0
\end{equation}
\begin{equation}\label{p1constraint2}
  \tilde{u}_{1,x}(R_{3}p_{1}+t_{z})-(R_{1}p_{1}+t_{x})=0
\end{equation}

Naturally, given another correspondence $u_2$ and $p_2$, according to $\{C,\tilde{u}_{2},{Rp_{2}+t}\}_{L}$
\begin{equation}\label{linep2}
  \frac{R_1 p_2 + t_x}{\tilde{u}_{2,x}} = \frac{R_2 p_2 + t_y}{\tilde{u}_{2,y}} = R_3 p_2 + t_z
\end{equation}
Then we can have two more constraints as
\begin{equation}\label{p2constraint1}
  \tilde{u}_{2,x}(R_{2}p_{2}+t_{y})-u_{2,y}(R_{1}p_{2}+t_{x})=0
\end{equation}
\begin{equation}\label{p2constraint2}
  \tilde{u}_{2,x}(R_{3}p_{2}+t_{z})-(R_{1}p_{2}+t_{x})=0
\end{equation}
Combining (\ref{p1constraint1}) - (\ref{p1constraint2}), $t_y$ and $t_z$ can be eliminated, then substituted into (\ref{p2constraint1}) - (\ref{p2constraint2}), $t_x$ can also be eliminated, resulting in an constraint only relating to $R$. Recall (\ref{eq.R}), by reorganizing the coefficients, we have the point-TIM as
\begin{equation}\label{tim}
  d_p(\alpha) = d_{p,1} \sin \alpha + d_{p,2} \cos \alpha + d_{p,3}
\end{equation}

\subsection{Derivation of line-TIM}

Each line feature correspondence can be represented by a pair of start point and end point of the line segment as shown in Fig.~\ref{fig.coordinates}. According to the projection geometry, the optical center of the camera, the 2D line segment $(u_{31},u_{32})$ and the 3D line $(p_{3},p_{4})$ lie on the same plane. Then the four points $C$, $u_{31}$, $u_{32}$ and $p_{3}$ are coplanar, denoted as $\{C,u_{31},u_{32},p_{3}\}_{P}$. Similarly, $\{C,u_{31},u_{32},p_{4}\}_{P}$ also holds. By solving the plane equation from the first three points and substituting the fourth point into it, we have:
\begin{equation}\label{sp1}
  (\tilde{u}_{31} \times \tilde{u}_{32})^T(Rp_3+t) = 0
\end{equation}
That is
\begin{align}\label{lineconstraints1}
  &(u_{31,y}-u_{32,y})(R_{1}p_{3}+t_{x})-(u_{31,x}-u_{32,x})(R_{2}p_{3}+t_{y})\nonumber\\
  &+(u_{31,x}u_{32,y}-u_{32,x}u_{31,y})(R_{3}p_{3}+t_{z})=0
\end{align}
Similarly, for $\{C,u_{31},u_{32},p_{4}\}_{P}$, we have:
\begin{equation}\label{sp2}
  (\tilde{u}_{31} \times \tilde{u}_{32})^T(Rp_4+t) = 0
\end{equation}
That is
\begin{align}\label{lineconstraints2}
  &(u_{31,y}-u_{32,y})(R_{1}p_{4}+t_{x})-(u_{31,x}-u_{32,x})(R_{2}p_{4}+t_{y})\nonumber\\
  &+(u_{31,x}u_{32,y}-u_{32,x}u_{31,y})(R_{3}p_{4}+t_{z})=0
\end{align}
With (\ref{lineconstraints1})-(\ref{lineconstraints2}), the $t$ can be eliminated resulting in
\begin{align}\label{lineconstraint}
  &[(u_{31,y}-u_{32,y})R_{1}-(u_{31,x}-u_{32,x})R_{2}\nonumber\\
  &+(u_{31,x}u_{32,y}-u_{32,x}u_{31,y})R_{3}](p_3-p_4)=0
\end{align}
Recall (\ref{eq.R}), (\ref{lineconstraint}) can be reorganized to line-TIM as
\begin{equation}\label{timl}
  d_l(\alpha) = d_{l,1} \sin \alpha + d_{l,2} \cos \alpha + d_{l,3}
\end{equation}

\subsection{Derivation of TIMs' lower bound}

Recall (\ref{tim}) and (\ref{timl}), as the forms of point-TIM and line-TIM are the same, we denote them as $d(\alpha)$. That is
\begin{equation}\label{tims}
\begin{split}
  d(\alpha) &= d_{1} \sin \alpha + d_{2} \cos \alpha + d_{3} \\
  &=\sqrt{d_1^2+d_2^2} (\sin \alpha \cos a_2 + \cos \alpha \sin a_2)+d_3 \\
  &=a_1 \sin (\alpha + a_2)+d_3
\end{split}
\end{equation}
where $a_1=\sqrt{d_1^2+d_2^2}$, $\sin a_2 = \frac{d_2}{a_1}$, $\cos a_2 = \frac{d_1}{a_1}$.

Then the lower bound of $|d(\alpha)|$ on $\mathbb{A}$, denoted as $\underline{d}(\mathbb{A})$, is derived as
\begin{equation}\label{timlb}
  \underline{d}(\mathbb{A}) =  \min |a_1 \sin(\alpha + a_2) + d_3|
\end{equation}

\section{Derivation of Translation Bound}

After the rotation estimation, we get the optimal yaw angle $\hat{\alpha}$. As shown in Fig.~\ref{fig.coordinates}, according to $\{C,\tilde{u}_{1},{R(\hat{\alpha})p_{1}+t}\}_{L}$, we have
\begin{equation}\label{pointL1}
  \tilde{u}_{1}\times(R(\hat{\alpha})p_{1}+t)=0
\end{equation}
which is equal to
\begin{equation}\label{pointL12}
  \tilde{u}_{1\times}(R(\hat{\alpha})p_{1}+t)=0
\end{equation}
where $a_{\times}$ denotes the symmetric matrix of vector $a$. Then (\ref{pointL12}) can be written as
\begin{equation}\label{pointL123}
  \begin{bmatrix}
    0 & -1 & \tilde{u}_{1,y} \\
    1 & 0 & -\tilde{u}_{1,x} \\
    -\tilde{u}_{1,y} & \tilde{u}_{1,x} & 0
  \end{bmatrix}
  \begin{bmatrix}
    t_x+h_1 \\
    t_y+h_2 \\
    t_z+h_3
  \end{bmatrix}=
  \begin{bmatrix}
    0 \\
    0 \\
    0
  \end{bmatrix}
\end{equation}
where $R(\hat{\alpha})p_{1}\triangleq(h_1,h_2,h_3)^T$. Then two equations of translation can be derived as
\begin{equation}\label{pointcon11}
  -t_y-h_2+\tilde{u}_{1,y}(t_z+h_3)=0
\end{equation}
\begin{equation}\label{pointcon12}
  t_x+h_1-\tilde{u}_{1,x}(t_z+h_3)=0
\end{equation}
Similarly, with another point correspondence $\{C,\tilde{u}_{2},{R(\hat{\alpha})p_{2}+t}\}_{L}$, we have
\begin{equation}\label{pointL2}
  \begin{bmatrix}
    0 & -1 & \tilde{u}_{2,y} \\
    1 & 0 & -\tilde{u}_{2,x} \\
    -\tilde{u}_{2,y} & \tilde{u}_{2,x} & 0
  \end{bmatrix}
  \begin{bmatrix}
    t_x+h_4 \\
    t_y+h_5 \\
    t_z+h_6
  \end{bmatrix}=
  \begin{bmatrix}
    0 \\
    0 \\
    0
  \end{bmatrix}
\end{equation}
where $R(\hat{\alpha})p_{2}\triangleq(h_4,h_5,h_6)^T$. Then we have another two equations as
\begin{equation}\label{pointcon21}
  -t_y-h_5+\tilde{u}_{2,y}(t_z+h_6)=0
\end{equation}
\begin{equation}\label{pointcon22}
  t_x+h_4-\tilde{u}_{2,x}(t_z+h_6)=0
\end{equation}
Combining (\ref{pointcon11})-(\ref{pointcon12}) and (\ref{pointcon21})-(\ref{pointcon22}), the translation can be solved as
\begin{eqnarray}\label{pointresult}
  t_x=\frac{\tilde{u}_{1,x}}{\tilde{u}_{1,y}-\tilde{u}_{2,y}}(\tilde{u}_{2,y}(h_6-h_3)+h_2-h_5)-h_1 \\
  t_y=\frac{\tilde{u}_{1,y}}{\tilde{u}_{1,y}-\tilde{u}_{2,y}}(\tilde{u}_{2,y}(h_6-h_3)+h_2-h_5)+h_2 \\
  t_z=\frac{1}{\tilde{u}_{1,y}-\tilde{u}_{2,y}}(h_2-h_5-\tilde{u}_{1,y}h_3+\tilde{u}_{2,y}h_6)
\end{eqnarray}

In addition, the translation can also be solved with one point and one line correspondence. According to (\ref{sp1})
\begin{equation}\label{line1}
  (\tilde{u}_{31}\times\tilde{u}_{32})^T(R(\hat{\alpha})p_3+t) = 0
\end{equation}
we have
\begin{equation}\label{linen}
\tilde{u}_{31} \times\tilde{u}_{32}=
\begin{bmatrix}
  \tilde{u}_{1,y}-\tilde{u}_{2,y} \\
  \tilde{u}_{1,x}+\tilde{u}_{2,x} \\
  \tilde{u}_{1,x}\tilde{u}_{2,y}-\tilde{u}_{2,x}\tilde{u}_{1,y}
\end{bmatrix}
\triangleq
\begin{bmatrix}
  n_1 \\
  n_2 \\
  n_3
\end{bmatrix}
\end{equation}
Then (\ref{sp1}) can be written as
\begin{equation}\label{line11}
  \begin{bmatrix}
    n_1 & n_2 & n_3
  \end{bmatrix}
  \begin{bmatrix}
    t_x+m_1 \\
    t_y+m_2 \\
    t_z+m_3
  \end{bmatrix}=0
\end{equation}
where $R(\hat{\alpha})p_3+t\triangleq(m_1,m_2,m_3)^T$. Similarly, with (\ref{sp2}), we have
\begin{equation}\label{line21}
  \begin{bmatrix}
    n_1 & n_2 & n_3
  \end{bmatrix}
  \begin{bmatrix}
    t_x+m_4 \\
    t_y+m_5 \\
    t_z+m_6
  \end{bmatrix}=0
\end{equation}
where $R(\hat{\alpha})p_4+t\triangleq(m_4,m_5,m_6)^T$. Thus, combining (\ref{pointcon11})-(\ref{pointcon12}) and (\ref{line11})-(\ref{line21}), the translation can be solved as
\begin{align}\label{lineresult}
  t_z=&(-n_1\tilde{u}_{1,x}h_3+n_1h_1-n_1m_1-n_2\tilde{u}_{1,y}h_3-n_2h_2 \nonumber\\
  &-n_2m_2-n_3m_3)/(n_1\tilde{u}_{1,x}+n_2\tilde{u}_{1,y}+n_3)
\end{align}
\begin{eqnarray}\label{lineresult2}
  t_y=\tilde{u}_{1,y}(t_z+h_3)+h_2 \\
  t_x=\tilde{u}_{1,x}(t_z+h_3)-h_1
\end{eqnarray}

Recall (\ref{3d2d}), there is \emph{unknown but bounded} \cite{milanese1989estimation} noise on the detected image features, such that $|e_i| < n_i$, and we have
\begin{equation}\label{featurebound}
  \underline{u}_i = u_{i}-n_i, \overline{u}_i = u_i+n_i
\end{equation}
With this feature bound of $u_{i}$, the un-normalized direction vector $\tilde{u_{i}}$ can also be bounded after linear transformations. Then the bound of the derived translation $[\ \underline{t}_{ij},\overline{t}_{ij}\ ]$ can be computed with the following relaxation \cite{mccormick1976computability} \cite{sherali1992new}:
\begin{equation}\label{eq.18}
\begin{split}
  f&=ab, \ \underline{a}\leq a \leq \overline{a}, \ \underline{b}\leq b \leq \overline{b}, \\
  f &\geq \max(\underline{a}b+\underline{b}a-\underline{a}\underline{b},\overline{a}b+\overline{b}a-\overline{a}\overline{b})\\
  f &\leq \min(\overline{a}b+\underline{b}a-\overline{a}\underline{b},\underline{a}b+\overline{b}a-\underline{a}\overline{b})
\end{split}
\end{equation}

\section{Real World Experiment Details}

The dataset employed in real world cross-session visual inertial localization task is YQ-dataset\cite{ding2018laser}. In the dataset, there are three sessions collected at summer 2017 in three days, denoted as 2017-0823, 2017-0827 and 2017-0828, and one session collected in winter 2018 after snow, denoted as 2018-0129. The 3D map is built with 2017-0823 session and the 3D-2D point feature correspondences are obtained by running visual inertial SLAM \cite{mur2017visual}. For evaluation, we compute the ground truth of the relative pose between the query camera and the map by aligning the synchronized LiDAR scans. For the pitch and roll angle, we use the estimation generated by visual inertial odometry \cite{mur2017visual}. To get the 3D-2D feature matches between the query image and the map, we exploited the following steps:
\begin{itemize}
\item Obtain the camera poses and the 3D-2D point matches in the map using visual inertial SLAM software \cite{mur2017visual}.
\item Run Line3D++ algorithm \cite{hofer2017efficient} to get the 3D-2D line matches in the map.
\item For the query session, we get the 3D-2D points/lines match based on the descriptors of LibVISO2 \cite{Geiger2011IV} and LBD \cite{zhang2013efficient}.
\end{itemize}

\end{appendices}

\bibliographystyle{ieeetr} 
\bibliography{appendix}

\end{document}